\documentclass[10pt]{article}
\usepackage[letterpaper]{geometry}
\usepackage{hicss}
\usepackage{times}
\usepackage[none]{hyphenat}
\usepackage{url}
\usepackage{latexsym}
\usepackage{graphicx}
\graphicspath{{images/}}
\usepackage[
    style=apa,
  ]{biblatex}
\usepackage{times}
\usepackage{latexsym}

\usepackage{graphicx}
\usepackage{subfig}
\usepackage{amsmath}
\usepackage{graphbox}
\usepackage{xcolor}
\definecolor{lavender}{RGB}{200, 195, 225}
\usepackage{microtype}

\title{The Reliability Paradox: Exploring How Shortcut Learning Undermines Language Model Calibration}

\author{Geetanjali Bihani \\
Purdue University \\
 {\underline{ gbihani@purdue.edu}} \\ \And
Julia Rayz\\
Purdue University  \\
 {\underline{ jtaylor1@purdue.edu} }
 }

\begin{document}
\maketitle
\begin{abstract}
The advent of pre-trained language models (PLMs) has enabled significant performance gains in the field of natural language processing. However, recent studies have found PLMs to suffer from miscalibration, indicating a lack of accuracy in the confidence estimates provided by these models. Current evaluation methods for PLM calibration often assume that lower calibration error estimates indicate more reliable predictions. However, fine-tuned PLMs often resort to shortcuts, leading to overconfident predictions that create the illusion of enhanced performance but lack generalizability in their decision rules. The relationship between PLM reliability, as measured by calibration error, and shortcut learning, has not been thoroughly explored thus far. This paper aims to investigate this relationship, studying whether lower calibration error implies reliable decision rules for a language model. Our findings reveal that models with seemingly superior calibration portray higher levels of non-generalizable decision rules. This challenges the prevailing notion that well-calibrated models are inherently reliable. Our study highlights the need to bridge the current gap between language model calibration and generalization objectives, urging the development of comprehensive frameworks to achieve truly robust and reliable language models.
\end{abstract}

\subsubsection*{Keywords:}

Pretrained language models, calibration, shortcut learning, robustness, generalization

\section{Introduction}

Pre-trained language models (PLMs) have become the convention in the field of natural language processing. The preference for PLMs can be attributed to their improvements in a wide variety of tasks, including question answering, textual entailment, sentiment analysis, and commonsense reasoning \parencite{peters2018deep, devlin_bert_2019, sap-etal-2020-commonsense}. The \textit{`pretrain-then-fine-tune'} paradigm allows the model to not only utilize the existing `knowledge' gained during pre-training but also learn from task-specific data via fine-tuning \parencite{alt_fine-tuning_2019, chen_bert_2019}.

Although fine-tuning PLMs achieves state-of-the-art results, it also causes models to lack generalization and become unreliable predictors. Specifically, PLMs tend to learn shortcuts based on keywords \parencite{moon_masker_2021, du-etal-2021-towards}  and cues related to language variations \parencite{nguyen_learning_2021} to make predictions. This behavior, also known as shortcut learning, leads the model to learn non-generalizable decision rules that do not perform well on out-of-distribution (OOD) data \parencite{moon_masker_2021, du2022shortcut}. Additionally, the fine-tuning process can lead to overconfidence in PLMs \parencite{kong_calibrated_2020, jiang_how_2021}, where their confidence increases regardless of the accuracy of their predictions \parencite{chen_close_2022}. This mismatch between the model's confidence and its actual accuracy in its predictions results in \textit{`miscalibration'} in language models.

It is desirable for language models to perform reliably and accurately across different language tasks. Addressing calibration is essential because it ensures that the model’s confidence aligns more accurately with its predictive accuracy. Miscalibrated models can lead to significant issues, particularly in high-stakes environments where wrong but confident predictions are dangerous. By focusing on calibration, we can reduce the mismatch between confidence and correctness, improving the model’s trustworthiness and robustness across diverse tasks and data distributions. Thus, it is important to study the interplay between model generalization and calibration. Prior works assessing model calibration focus on measuring and minimizing statistical calibration evaluation metrics such as Expected Calibration Error (ECE) \parencite{kong_calibrated_2020, ahuja_calibration_2022, kim_bag_2023}. These works do not investigate whether lower calibration error estimates align with more generalizable decision rules learned by these language models.

In this study, we aim to address this gap and conduct the following research inquiries: 1) Does a reduction in calibration error within language models indicate a decrease in overconfident predictions? 2) Can a model exhibiting lower calibration error be considered reliable in terms of its decision rules? By examining these questions, we seek to shed light on the relationship between calibration error and the reliability of language models' decision rules. Our questions are based on the intuition that model reliability estimates should account for the reliability of the model's decision rules. 

To answer these questions, we investigate the calibration and shortcut learning behaviors of recent pre-trained language models (PLMs) across a suite of binary and multi-class classification tasks, and analyze the evolution of shortcut learning behaviors in PLMs before and after fine-tuning.

Our research findings highlight that models appearing to be well-calibrated often exhibit a higher propensity for shortcut learning. This challenges the conventional notion of well-calibrated models as reliable and robust. While lower calibration error estimates, such as Expected Calibration Error (ECE), may indicate the improved alignment of prediction probabilities with their actual correctness, they fail to capture the inherent lack of robustness in these `correct' model decisions. This observation uncovers a fundamental discrepancy between the requirements of model calibration and the goals of generalization, highlighting the need to reconcile these seemingly contradictory frameworks in order to achieve truly robust and reliable language models.

\noindent{\textbf{Contributions:}}
Our contributions are summarized as follows.
\begin{itemize}
    \item Our study shows that contrary to previous assumptions, a lower Expected Calibration Error (ECE) does not necessarily indicate improved reliability. Instead, it often reflects models' tendency to make overconfident predictions driven by shortcut cues. 
    \item We perform analyses on fine-tuned PLMs, across a suite of text classification tasks, highlighting the limitations of statistical calibration error measures such as ECE, in capturing the lack of robustness in model decisions. This insight underscores the need to consider the trade-off between task performance and robustness to shortcuts when evaluating model calibration.
\end{itemize}

\section{Shortcut Effects on Calibration}
\subsection{Identifying Shortcuts}\label{sec:2.1}

Shortcut learning refers to the phenomenon where models rely on superficial cues in the training data to make predictions instead of learning the underlying semantics to perform an NLU task. This over-reliance on specific features or biases results in poor generalization in out-of-distribution (OOD) settings. Identifying shortcut learning in language models is an ongoing research area, with recent works utilizing model attention, dataset statistics, and human annotated samples to identify spurious correlations \parencite{moon_masker_2021, wang-etal-2022-identifying}. We utilize the shortcut identification framework as described by \parencite{du-etal-2021-towards}, which combines data statistics with model attributions to identify shortcuts. We describe this shortcut identification framework below.

\noindent{\textbf{Model Attribution based Importance: }} To obtain attributions for each token ($w_{j}$) in a given sample $S_{i}$,  we utilize integrated gradients (IG) \parencite{sundararajan2017axiomatic}. Let a given sample $S_{i}$ contain $T$ tokens, i.e. $S_{i} = \left\{w_{j}^{t}\right\}_{t=1}^{T}$. We conduct a step-wise perturbation of the sample, creating $m$ intermediate samples along a straight-line path from a baseline $S_{b}$ to the actual sample $S_{i}$. By observing the changes in the model's output as the sample is progressively modified, we quantify the contribution of each token to the final prediction. Following \parencite{du-etal-2021-towards}, we consider all-zero embeddings to form $S_{b}$. As each word is added to the baseline, the gradient of the prediction $M(S_{i}$) is computed  with respect to the associated token embeddings ($e(w_{j})$) obtained from the output embedding layer of model $M$. The following equations summarize this gradient calculation.

\begin{equation}\label{eq:part2}
IG\left(S_{i}\right) = S^{\text{b}}_{i}\cdot\sum_{k=1}^{m} \frac{\partial M_{y}\left(S_{\text{b}}+\frac{k}{m}\left(S_{i}-S_{\text{b}}\right)\right)}{\partial S_{i}} \cdot \frac{1}{m}
\end{equation}
where
\begin{equation}\label{eq:part1}
S^{\text{b}}_{i} = S_{i}-S_{\text{b}}
\end{equation}
Finally, the L2 norm between the gradient and the corresponding token embedding is calculated to determine the individual contribution of each token. Following \parencite{du-etal-2021-towards}, we filter top three attributed tokens per sample.

\noindent{\textbf{Local Mutual Information based Importance: }}  We calculate local mutual information (LMI) between tokens and task labels for a given dataset $D$, using Eq.\ref{eq:lmi}. 
\begin{equation}\label{eq:lmi}
\operatorname{LMI}(w, y)=p(w, y) \cdot \log \left(\frac{p(y \mid w)}{p(y)}\right)
\end{equation}
where $p(w, y) = \frac{count(w,y)}{\left|V\right|}$, $p(y \mid w) = \frac{count(w,y)}{count(w)}$ and $p(y) = \frac{count(y)}{\left|V\right|}$. Here $w$ refers to a word token appearing in the samples with the task label $y$, and $\left|V\right|$ refers to the size of the vocabulary of dataset $D$.

\noindent{\textbf{Comparing Importance to Estimate Shortcuts: }} As mentioned in \parencite{du-etal-2021-towards}, for a given label $y$, we consider the top $5\%$ LMI-scored tokens as the \textit{head} of the label's LMI distribution. For a given sample, if the top-$3$ attributed tokens for a prediction also appear in the \textit{head} of the predicted label's LMI distribution, it is termed as a shortcut.  

\subsection{Types of Shortcuts}\label{sec:2.2}

\begin{table}
\centering
\begin{tabular}{lll}
\hline \textbf{Text} & \textbf{Label} & \textbf{Shortcuts}  \\ \hline
This is awesome!  & \textit{positive} & `!', `awesome'  \\
This is tragic... & \textit{negative} & `...', `tragic' \\ \cline{1-2}
\hline
\end{tabular}
\caption{Example shortcuts in binary sentiment classification}\label{sentiment-example} 
\end{table}

\begin{table*}
\centering
\begin{tabular}{lp{0.6\linewidth}l}
\hline \textbf{Type} & \textbf{Text} & \textbf{Data} \\ \hline
\textit{Lexicon-cued} & \colorbox{lavender}{Michael} \colorbox{lavender}{Phelps} \colorbox{lavender}{won} the gold medal in the 400 individual medley and set a world record in a time of 4 minutes 8.26 seconds. & AG News\\
\hline
\textit{Grammar-cued} & What are spider veins \colorbox{lavender}{?} & TREC\\
\textit{} & Guess \colorbox{lavender}{they} didn't get the memo reg non-nuclear Baltic sea \#sarcasm & TweetEval (Irony)\\
\hline
\end{tabular}
\caption{Examples of shortcuts learned by PLMs across different tasks}\label{phelps-example} 
\end{table*}

Within the broader context of language structure, a well-established theoretical framework proposed by \parencite{chomsky1965aspects} introduced a fundamental division between the lexicon and grammar. According to this framework, the lexicon serves as a repository for language words, while grammar establishes rules for combining these words. Drawing upon these foundational concepts, we divide the identified shortcuts into two categories, i.e. \emph{`lexicon-cued'} and \emph{`grammar-cued'} predictions. PLM predictions where at least one lexical word is utilized are classified as \emph{`lexicon-cued'} predictions. On the other hand, predictions where the identified shortcuts are limited to functional words and punctuations, are labeled as \emph{`grammar-cued'} predictions. The purpose of this categorization is to improve our understanding of the shortcut mechanisms employed by the models in their prediction processes, particularly in terms of their lexical-semantic processing. Examples are shown in Table~\ref{phelps-example}

For a more fine-grained analysis, we differentiate between cases where a model exclusively relies on punctuation, stopwords, or sub-words for making predictions and cases where it additionally incorporates one or more lexical words. This distinction allows us to compare the model's reliance on grammatical and lexical cues. For example, considering the case given in Table \ref{sentiment-example}, if a sentiment classification model relies on an exclamation mark \textit{`!'} to predict the sentiment of \textit{`this is awesome!'}, the rule is less generalizable as compared to relying on the word \textit{`awesome'}. This is because punctuations and stopwords lack explicit semantic information on sentiment and can be used in various contexts to indicate surprise, exparencitement, or emphasis. On the other hand, the word \textit{`awesome'} has a more consistent and explicit semantic association with positive sentiment and is frequently used in positive contexts, and has a positive polarity (sarcasm excluded). 

\subsection{Measuring Calibration}\label{sec:2.3}
A well-calibrated model should provide accurate probability estimates that reflect the true likelihood of an event. To quantify model calibration, we measure the Expected Calibration Error (ECE) \parencite{naeini2015obtaining}. We choose ECE because it captures the discrepancy between the model’s confidence and accuracy, and has been used for calibration analysis, making it an ideal choice for evaluating the model's reliability across a range of tasks. This metric calculates the weighted average of the difference between the accuracy of a model and its average confidence level over a set of bins defined by the predicted probabilities, as shown in Eq. \ref{ece}, where $n$ is the number of samples in $B_{m}$.

\begin{equation}\label{ece}
\mathrm{ECE}=\sum_{m=1}^{M} \frac{\left\vert B_{m}\right\vert}{n} \mid \operatorname{acc}\left(B_{m}\right)-\operatorname{conf}\left(B_{m}\right)
\end{equation}

Here, the estimation of expected accuracy from finite samples is done by grouping predictions ($\hat{p}_{i}$) into $M$ interval bins (each of size $\frac{1}{M}$), and the accuracy of each bin is calculated. Let $B_{m}$ be a bin containing samples whose prediction confidence lies within the interval $I_{m}=\left(\frac{m-1}{M}, \frac{m}{M}\right]$. Then the accuracy of $B_{m}$, where $y_{i}$ and $\hat{y}_{i}$ portray predicted and true class labels, is calculated as shown in Eq. \ref{acc_cal}.

\begin{equation}\label{acc_cal}
\operatorname{acc}\left(B_{m}\right)=\frac{1}{\left\vert B_{m} \right\vert}\sum_{i \in B_{m}} \mathbf{1}\left(\hat{y}_{i}=y_{i}\right)
\end{equation}

The average predicted confidence of $B_{m}$, is calculated as shown in Eq. \ref{conf_cal}.

\begin{equation}\label{conf_cal}
\operatorname{conf}\left(B_{m}\right)=\frac{1}{\left\vert B_{m} \right\vert}\sum_{i \in B_{m}} \hat{p}_{i}
\end{equation}

\subsection{Measuring trade-offs}
In order to investigate the relationship between model calibration and shortcut learning, we calculate two metrics: the portion of shortcut-cued model predictions ($P_{sc}$) and the shortcut trade-off ($T_{sc}$). The calculation of $P_{sc}$ allows us to quantify the extent to which a model relies on shortcuts when making predictions. Additionally, we introduce $T_{sc}$ as a metric to assess the trade-off between shortcut learning and model performance. $T_{sc}$ is calculated as the ratio of task accuracy (e.g., F1 score) to the proportion of shortcut-cued predictions. 

\begin{equation}
T_{sc} = \frac{\text{Task Accuracy } (F1)}{\text{Shortcut-Cued Predictions } (P_{sc})}
\end{equation}

A higher $T_{sc}$ score indicates that the model achieves better task accuracy while relying on fewer shortcut-cued predictions. Conversely, a lower $T_{sc}$ score suggests a higher reliance on shortcuts to achieve optimal task performance. In our analysis, we aim to maximize $T_{sc}$ and minimize Expected Calibration Error (ECE) in order to identify models that strike a balance between shortcut learning and accurate predictions.

\begin{table*}
\centering
\begin{tabular}{lcccc}

\hline
$P_{sc}$ / $T_{sc}$ / ECE & \textbf{COLA}                    & \textbf{Hate}                    & \textbf{Irony}                   & \textbf{SST2}                    \\
\hline
{ALBERT}  & 51.29 / 1.66 / 0.16 & 91.72 / 0.67 / 0.42 & 79.97 / 0.72 / 0.29 & 84.98 / 1.06 / 0.44 \\
{BART}     & 53.98 / 1.58 / 0.22 & 94.28 / 0.65 / 0.47 & 72.19 / 0.83 / 0.31 & 83.83 / 1.12 / 0.45 \\
{BERT}    & 66.73 / 1.28 / 0.11 & 95.39 / 0.65 / 0.36 & 71.68 / 0.55 / 0.15 & 88.88 / 1.03 / 0.44 \\
{DeBERTa} & 54.07 / 1.63 / 0.22 & 95.35 / 0.66 / 0.48 & 86.10 / 0.75 / 0.35  & 84.86 / 1.12 / 0.47 \\
{RoBERTa} & 59.16 / 1.45 / 0.15 & 91.55 / 0.68 / 0.37 & 83.80 / 0.40 / 0.14   & 84.52 / 1.11 / 0.45 \\
\hline
$P_{sc}$ / $T_{sc}$ / ECE & \textbf{AG News}                 & \textbf{Emotion}                 & \textbf{Sentiment}               & \textbf{TREC}                    \\
\hline
{ALBERT}  & 92.11 / 1.02 / 0.01 & 79.38 / 0.64 / 0.08 & 84.79 / 0.79 / 0.04 & 93.60 / 0.95 / 0.05 \\
{BART}    & 95.93 / 0.99 / 0.01 & 68.47 / 1.10 / 0.04  & 76.89 / 0.93 / 0.06 & 91.20 / 1.03 / 0.02 \\
{BERT}    & 97.20 / 0.97 / 0.01  & 77.62 / 0.39 / 0.09 & 80.49 / 0.87 / 0.02 & 97.40 / 0.75 / 0.39 \\
{DeBERTa} & 96.50 / 0.98 / 0.01  & 75.09 / 1.03 / 0.03 & 75.29 / 0.94 / 0.09 & 95.20 / 0.99 / 0.03 \\
{RoBERTa} & 95.47 / 0.99 / 0.01 & 71.36 / 0.62 / 0.11 & 72.92 / 0.98 / 0.03 & 84.52 / 0.82 / 0.11 \\
\hline
\end{tabular}%
\caption{Table comparing PLMs across shortcuts learnt ($P_{sc}$), shortcut trade-off ($T_{sc}$) and calibration (ECE). Top row: COLA, Hate, Irony and SST2 are binary classification tasks; Bottom Row: AG News, Emotion, Sentiment and TREC are multi-class classification tasks.}\label{f1-ece-shortcuts-here}
\end{table*}

\section{Experiments}

\noindent{\textbf{Datasets.}}
To evaluate PLM shortcut learning and calibration effects across different tasks and domains, we perform our evaluation on several binary and multi-class classification tasks. Specifically, we consider 8 text classification datasets, briefly described as follows: i) Stanford-Sentiment Treeback (SST-2) \parencite{socher2013}, commonly used in sentiment analysis tasks and provides a valuable benchmark for evaluating models` ability to capture sentiment in texts, ii) Corpus of Linguistic Acceptability (COLA) \parencite{warstadt2019neural}, which assesses model performance on grammaticality judgments, iii) TREC (coarse-grained) \parencite{hovy-etal-2001-toward} used for question classification, iv) AG News \parencite{Zhang2015CharacterlevelCN} used for news topic classification, and four datasets from TweetEval benchmark \parencite{barbieri2020} [Emotion, Hate, Irony, and Sentiment] for text classification in the context of short and informal social media texts. These datasets present different challenges, such as the presence of sarcasm and negation in samples for sentiment tasks, lexical overlap in topic classification tasks, and the inclusion of short and ironic social media texts in irony detection tasks. This language and task variation across datasets allows for a holistic assessment of the models' performance and generalization of our findings in the context of PLM calibration literature.

\noindent{\textbf{Models.}} We evaluate five pre-trained transformer language models for evaluation: BERT \parencite{devlin_bert_2019}, RoBERTa \parencite{liu2019roberta}, DeBERTa \parencite{hedeberta}, ALBERT \parencite{lanalbert} and BART \parencite{lewis2020bart}. We choose BERT and RoBERTa to align our results with prior PLM calibration research \parencite{desai2020calibration, kim_bag_2023}. We additionally evaluate more recent transformer LMs including DeBERTa and ALBERT. DeBERTa improves model generalization on downstream tasks, compared to BERT and RoBERTa, attributed to its disentangled attention mechanism. ALBERT is a compact architecture, providing performance gains with minimal sacrifice of task performance. Finally, we include BART, due to its improvements in handling of global context and robustness in handling noisy and ambiguous texts.

\begin{figure}
     \centering
     \subfloat[Lexical-cued predictions]{\includegraphics[width=0.36\textwidth]{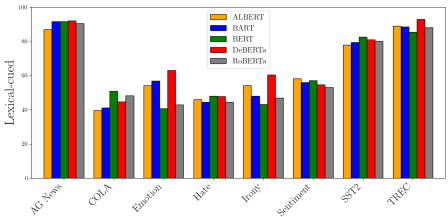}}
     \vfill
     \subfloat[Grammar-cued predictions]{\includegraphics[width=0.36\textwidth]{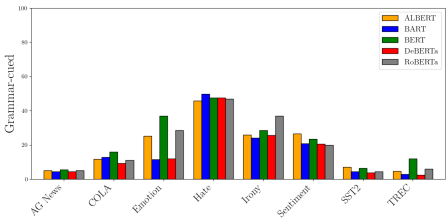}}
    \caption{Difference in distribution of shortcut-cued predictions across different tasks.}
    \label{fig:shortcuts-diff}
\end{figure}

\noindent{\textbf{Metrics.}} 
To investigate the association between shortcut learning effects and model calibration, we employ multiple evaluation metrics. We utilize F1 score to assess the overall prediction performance of PLMs on given tasks. Additionally, we measure the Expected Calibration Error (ECE) \parencite{naeini2015obtaining}, as described in Section \ref{sec:2.3}.

We also examine the distribution of shortcuts across correct and incorrect predictions using $P_{sc}$ and $T_{sc}$. These measures allow us to gain insights into the relationship between shortcut utilization and prediction accuracy. By analyzing how shortcuts are distributed across different prediction outcomes, we can explore their impact on the model's ability to classify accurately.

\noindent{\textbf{Training Configurations.}} 
Following prior work \parencite{kim_bag_2023}, we fix several hyperparameters for the model fine-tuning process. For all models, we set the initial learning rate to 1e-5 and gradient clip to 1.0. We utilized an Adam optimizer with an $\epsilon$ value of 1e-8, and set the batch size to 32. We fine-tuned our models over a maximum of 3 epochs. To gauge the impact of fine-tuning on shortcut learning and model calibration, we also use PLMs off-the-shelf to make predictions. We build our text classifiers using the Huggingface Transformers library\footnote{https://huggingface.co/models}. We report results averaged across five runs per task.

\section{Results \& Discussion}

\noindent{\textbf{Comparison Across Models and Tasks:}}
For all models in our analyses, we find more than chance ($>50\%$) shortcut learning for every task. Shown in Table \ref{f1-ece-shortcuts-here}, we observe a negative relationship between model calibration and shortcut trade-off ($T_{sc}$), i.e models considered more calibrated in terms of expected calibration error also tend to rely more on shortcuts when making predictions. This finding highlights that metrics like Expected Calibration Error (ECE), which assess statistical model calibration, do not align with the model's robustness in terms of learning fewer spurious correlations. Across models, we find that BERT and RoBERTa rely on more shortcut-cued predictions but appear statistically more calibrated. In contrast, DeBERTa and BART rely on fewer shortcuts but appear statistically less calibrated. 

\begin{figure*}
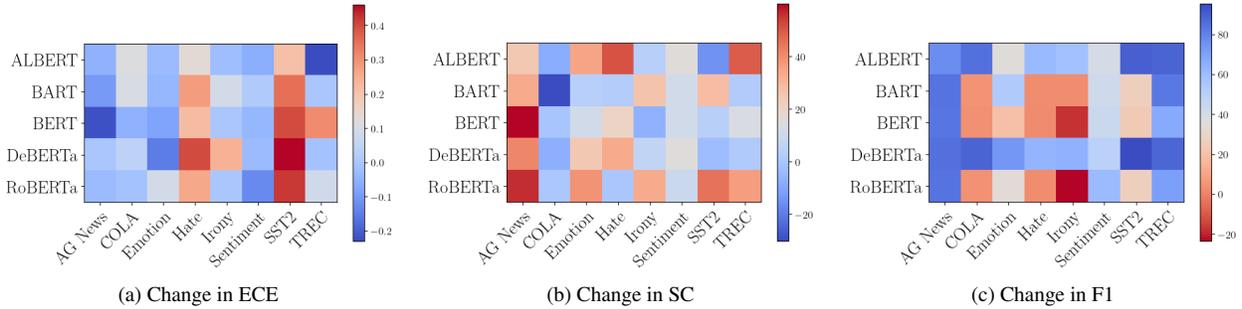

     \centering
    \subfloat[Change in ECE]{\includegraphics[width=0.32\textwidth]{images/ece_diff.pdf}}
     \hfill
     \subfloat[Change in SC]{\includegraphics[width=0.32\textwidth]{images/sc_diff.pdf}}
     \hfill
     \subfloat[Change in F1]{\includegraphics[width=0.32\textwidth]{images/f1_diff.pdf}}
    \caption{Change in model performance and calibration before and after fine-tuning. (a): red represents an increase and blue represents a decrease in ECE after fine-tuning, (b): red represents an increase and blue represents a decrease in shortcut-cued predictions, (c): red represents a decrease and blue represents an increase in F1 after fine-tuning}
        \label{fig:finetune_before_after}
\end{figure*}

\noindent{\textbf{Shortcuts Learned:}}
Across various datasets, we observe a notable difference in the extent of shortcut learning. Models make more lexicon-cued predictions on datasets such as AG News, SST2 and TREC, while more grammar-cued predictions are made on Hate and Irony datasets, as shown in Figure \ref{fig:shortcuts-diff}. Further, the results in Table \ref{f1-ece-shortcuts-here} show that models exhibit a higher degree of shortcut learning on AG News, Hate, and TREC, while COLA shows relatively lower levels. This difference can be attributed to the linguistic characteristics present in the task samples. In COLA, the data includes acceptability labels, and the word tokens appear in a wider range of contexts, not limited to specific topics or sentence formation (e.g. questions) for two distinct categories. On the other hand, AG News, Hate, and TREC samples contain more prevalent lexical cues that repeat across samples. Across AG News and TREC, while different types of question words consistently appear across respective question types in TREC, AG News contains topic-specific words like \textit{`President'} and \textit{`Minister'} appearing in the `World' category. Although these words can act as genuine cues, PLMs heavily rely on them, resulting in misclassification. Examples of this behavior are shown in Table \ref{missed-class}, where the highlighted tokens in given text examples are shortcuts typically associated with other labels.

\noindent{\textbf{Fine-tuning Effects:}}
We evaluate changes in shortcut learning ($P_{sc}$), task performance (F1), and calibration (ECE) in models due to fine-tuning. Figure \ref{fig:finetune_before_after} shows the changes observed in classification performance before and after fine-tuning PLMs. We observe that fine-tuning does not always lead to calibration improvements, which aligns with prior findings \parencite{kong_calibrated_2020, jiang_how_2021}. Note that models become increasingly miscalibrated on Hate and SST2 tasks, while showing improved calibration for AG News, attributed to the increased accuracy due to shortcut learning of models on AG News. We also find that shortcut learning reduces after fine-tuning for some models, especially on COLA, which we attribute to the words appearing across a wider variety of contexts in its samples, as the dataset is not constricted to specific topics or affect statements.

\begin{table*}
\centering
\begin{tabular}{lp{0.6\linewidth}ll}
\hline
\textbf{Data}   & \textbf{Text}    & \textbf{Actual}                        & \textbf{Predicted}                                         \\ \hline
AG News & U.S. Seeks Reconciliation with \colorbox{lavender}{Oil}-Rich Venezuela  SAO PAULO, Brazil (Reuters) - The United States said on  \colorbox{lavender}{Monday} it will seek better ties with oil-rich Venezuela in the  clearest sign since President Hugo Chavez won a recall referendum in August that \colorbox{lavender}{Washington} is looking for  reconciliation with the firebrand populist. & World     & Business               \\
\hline
TREC & \colorbox{lavender}{What} do bats eat \colorbox{lavender}{?} & Entity & Description \\
\hline   
\end{tabular}
\caption{Examples of misclassification due to shortcuts}\label{missed-class} 
\end{table*}

\begin{figure*}
     \centering 
     \subfloat[]{{\includegraphics[width=0.25\textwidth, align=c]{images/reliability-trec-deberta.pdf}}}
     \hfill
    \subfloat[]{{\includegraphics[width=0.25\textwidth, align=c]{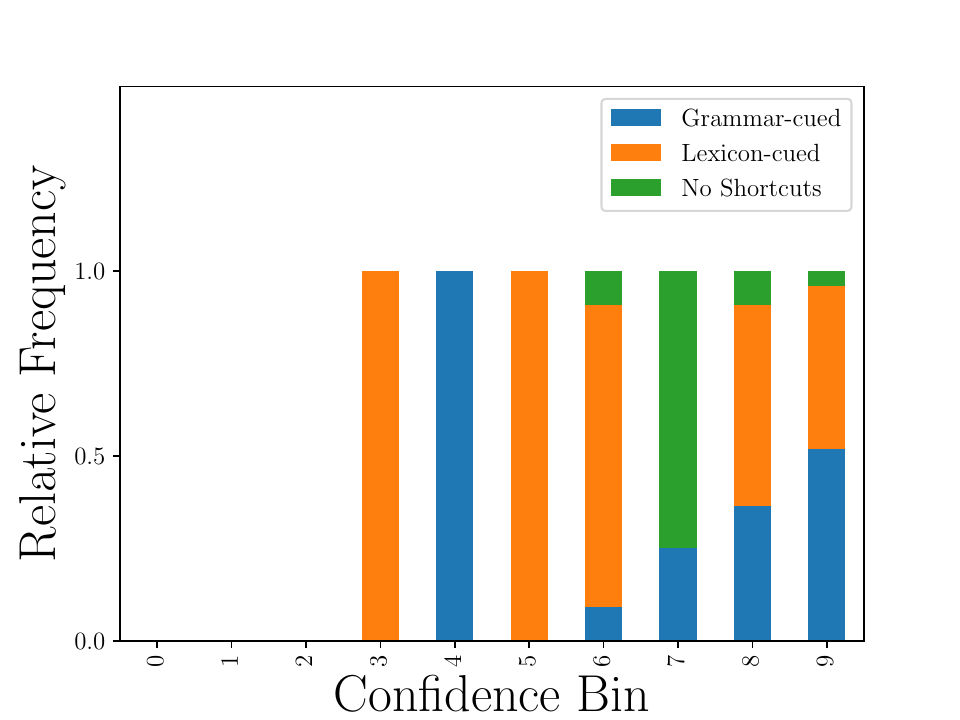}}}
    \hfill
    \subfloat[]{{\includegraphics[width=0.25\textwidth,align=c]{images/reliability-agnews-deberta.pdf}}}
    \hfill
    \subfloat[]{{\includegraphics[width=0.25\textwidth,align=c]{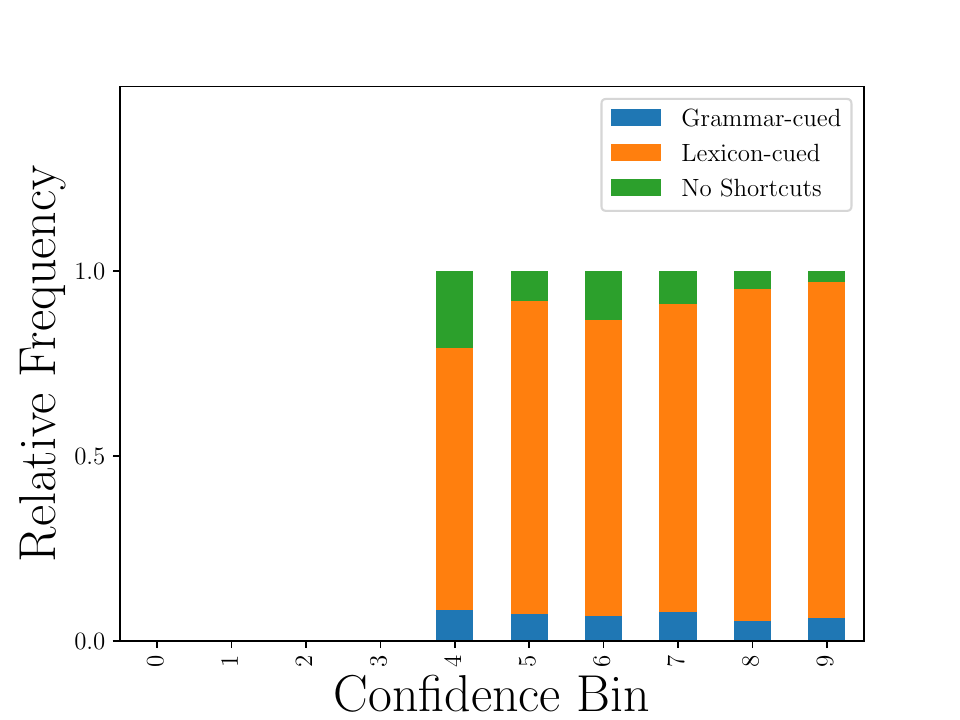}}}
    \caption{Difference in distribution of shortcut-cued predictions on fine-tuned DeBERTa for (a) TREC and (b) AG News. Models show similar performance on both tasks in terms of F1 and ECE; $F_1^{\text{AG News}} = 94.99$, $F_1^{\text{TREC}} = 94.06$; $ECE^{\text{AG News}} = 0.01$, $ECE^{\text{TREC}} = 0.03$.}
    \label{fig:shortcuts-diff-deberta}
\end{figure*}

\noindent{\textbf{Shortcut Impacts on Model Confidence:}} 
While fine-tuning models results in increased confidence on correct as well as incorrect predictions, we focus on instances where the calibration error estimates such as ECE are unable to capture the underconfident predictions, i.e. predictions that are correct, but less confident than average. In Figure \ref{fig:shortcuts-diff-deberta}, we plot reliability diagrams and shortcut-cued prediction distributions across model confidence for DeBERTA on two tasks, i.e. TREC and AG News. While DeBERTa portrays similar F1 and ECE for both tasks, we observe that the confidence and shortcut distributions are starkly different. Specifically, DeBERTa predictions on TREC are underconfident in many cases, while on AG News, the model confidence aligns with the correctness of model decisions. Further, while DeBERTa heavily relies on shortcuts for both the tasks, TREC relies more on grammar-cues, while AG News predictions are more often lexicon-cued. This discrepancy in model confidence and shortcuts utilized per confidence bin is not captured in statistical calibration error metrics like ECE.
It is crucial to highlight that statistical calibration error metrics like ECE fail to capture the divergence in model confidence and the specific shortcuts employed within confidence bins. While ECE may provide an overall assessment of model calibration, it falls short in capturing the complex interplay between confidence, shortcuts, and their impact on prediction reliability.

\noindent{\textbf{Is minimizing ECE enough?}}
We discover that ECE is not a dependable metric, and can be low even when the model is highly overconfident. Thus, a lower ECE does not necessarily indicate more reliable predictions by a model. To illustrate this, let's consider the results presented in Table \ref{f1-ece-shortcuts-here}, specifically for the Hate and AG News tasks. Both tasks demonstrate significant levels of shortcut learning in language models. However, the shortcuts learned in the AG News task lead to more accurate predictions, whereas shortcuts learned in the Hate task result in more incorrect predictions, inflating the latter's Expected Calibration Error (ECE) and causing it to be considered less calibrated. Note that a model fine-tuned on the AG News task may exhibit the appearance of being a reliable predictor, yet its reliability is inflated by the correctness of its predictions. For instance, in Figure \ref{fig:shortcuts-diff-deberta}, we find that underconfident model predictions are accurate in many cases. The ECE metric does not penalize accurate but underconfident and accurate predictions. 

This limitation underscores a significant drawback associated with using Expected Calibration Error (ECE) as the prevailing calibration estimate in PLM calibration literature. The use of ECE fails to consider the presence of spurious associations learned by language models.

\section{Related Work}
\noindent{\textbf{Shortcut Learning}}
General-purpose neural language models have been shown to learn spurious patterns existing within natural language text, due to the language variety cues within the training corpora \parencite{nguyen_learning_2021}. While initial research claimed that pre-trained language models are robust to out-of-distribution (OOD) detection and cross-domain generalization \parencite{hendrycks_pretrained_2020}, recent analyses have shown that PLMs, and their fine-tuned versions rely on specific keyword-based shortcuts to perform classification  \parencite{moon_masker_2021}. This phenomenon hinders the fine-tuned models from learning generalizable decision rules. PLMs have also been shown to rely on syntactic heuristics to perform natural language inference tasks \parencite{mccoy-etal-2019-right}. In light of these findings, research focusing on the automatic identification and mitigation of spurious cues within training and fine-tuning data has also been proposed \parencite{wang-culotta-2020-identifying, tu2020empirical, wang-etal-2022-identifying}.


\noindent\textbf{Calibration in Neural Language Models}
With the increased application of neural network architectures in high-risk real-world settings. their calibration has become an extensively studied topic in recent years \parencite{thulasidasan_mixup_2019, malinin_predictive_2018, hendrycks_pretrained_2020}. Recent research has focused on improving the calibration of neural networks, particularly in the context of deep learning. Various methods have been proposed to achieve better calibration, including temperature scaling \parencite{guo_calibration_2017}, isotonic regression \parencite{platt1999probabilistic}, and histogram binning \parencite{zadrozny2001obtaining}.

Pre-trained language models have garnered attention due to their tendency to exhibit increasing confidence during training, regardless of the accuracy of their predictions \parencite{chen_close_2022}. However, these models demonstrate better calibration within in-domain (ID) settings while experiencing calibration deterioration in out-of-domain (OOD) scenarios \parencite{desai_calibration_2020}. Interestingly, it has been observed that smaller models achieve improved calibration on in-domain data, whereas larger models exhibit superior calibration on out-of-domain data \parencite{dan_effects_2021}. 

Moreover, fine-tuning pre-trained language models leads to higher levels of miscalibration \parencite{kong_calibrated_2020, jiang_how_2021}. This is attributed to the excessive parameterization of the models, resulting in overfitting the training data. 
These findings highlight the inadequacy of current PLMs in terms of confidence calibration and reliability in decisions. 

\section{Conclusion}
The prevailing belief in existing calibration evaluations of pre-trained language models is that lower calibration error estimates indicate more reliable predictions. However, it has been shown that fine-tuned PLMs often rely on shortcuts to produce overly confident predictions, creating an illusion of improved performance while actually learning decision rules that lack generalizability. The relationship between model reliability, as measured by calibration error, and shortcut learning has received limited attention thus far. This prompts us to question whether a model with lower calibration error can truly be considered reliable in terms of its decision rules. Our findings challenge the prevailing notion by revealing that models with seemingly better calibration also exhibit higher levels of shortcut learning. This highlights the need to bridge the current gap between language model calibration and generalization objectives and underscores the importance of developing comprehensive frameworks to achieve genuinely robust and reliable language models.




\printbibliography

@article{du2022shortcut,
  title={Shortcut learning of large language models in natural language understanding: A survey},
  author={Du, Mengnan and He, Fengxiang and Zou, Na and Tao, Dacheng and Hu, Xia},
  journal={arXiv preprint arXiv:2208.11857},
  year={2022}
}

@inproceedings{mccoy-etal-2019-right,
    title = "Right for the Wrong Reasons: Diagnosing Syntactic Heuristics in Natural Language Inference",
    author = "McCoy, Tom  and
      Pavlick, Ellie  and
      Linzen, Tal",
    booktitle = "Proceedings of the 57th Annual Meeting of the Association for Computational Linguistics",
    month = jul,
    year = "2019",
    address = "Florence, Italy",
    publisher = "Association for Computational Linguistics",
    url = "https://aclanthology.org/P19-1334",
    doi = "10.18653/v1/P19-1334",
    pages = "3428--3448",
}

@inproceedings{desai2020calibration,
  title={Calibration of Pre-trained Transformers},
  author={Desai, Shrey and Durrett, Greg},
  booktitle={Proceedings of the 2020 Conference on Empirical Methods in Natural Language Processing (EMNLP)},
  pages={295--302},
  year={2020}
}

@book{chomsky1965aspects,
  title={Aspects of the theory of syntax},
  author={Chomsky, Noam},
  year={1965},
  publisher={MIT Press}
}

@article{liu2019roberta,
  title={Roberta: A robustly optimized bert pretraining approach},
  author={Liu, Yinhan and Ott, Myle and Goyal, Naman and Du, Jingfei and Joshi, Mandar and Chen, Danqi and Levy, Omer and Lewis, Mike and Zettlemoyer, Luke and Stoyanov, Veselin},
  journal={arXiv preprint arXiv:1907.11692},
  year={2019}
}

@inproceedings{nguyen_learning_2021,
	address = {Online},
	title = {On learning and representing social meaning in {NLP}: a sociolinguistic perspective},
	shorttitle = {On learning and representing social meaning in {NLP}},
	url = {https://aclanthology.org/2021.naacl-main.50},
	doi = {10.18653/v1/2021.naacl-main.50},
	urldate = {2021-08-24},
	booktitle = {Proceedings of the 2021 {Conference} of the {North} {American} {Chapter} of the {Association} for {Computational} {Linguistics}: {Human} {Language} {Technologies}},
	publisher = {Association for Computational Linguistics},
	author = {Nguyen, Dong and Rosseel, Laura and Grieve, Jack},
	month = jun,
	year = {2021},
	pages = {603--612},
}

@article{tu2020empirical,
  title={An empirical study on robustness to spurious correlations using pre-trained language models},
  author={Tu, Lifu and Lalwani, Garima and Gella, Spandana and He, He},
  journal={Transactions of the Association for Computational Linguistics},
  volume={8},
  pages={621--633},
  year={2020},
  publisher={MIT Press}
}

@article{jiang_how_2021,
	title = {How {Can} {We} {Know} {When} {Language} {Models} {Know}? {On} the {Calibration} of {Language} {Models} for {Question} {Answering}},
	volume = {9},
	issn = {2307-387X},
	shorttitle = {How {Can} {We} {Know} {When} {Language} {Models} {Know}?},
	url = {https://doi.org/10.1162/tacl_a_00407},
	doi = {10.1162/tacl_a_00407},
	urldate = {2023-03-06},
	journal = {Transactions of the Association for Computational Linguistics},
	author = {Jiang, Zhengbao and Araki, Jun and Ding, Haibo and Neubig, Graham},
	month = sep,
	year = {2021},
	pages = {962--977},
}

@inproceedings{ahuja_calibration_2022,
	address = {Abu Dhabi, United Arab Emirates},
	title = {On the {Calibration} of {Massively} {Multilingual} {Language} {Models}},
	url = {https://aclanthology.org/2022.emnlp-main.290},
	urldate = {2023-03-20},
	booktitle = {Proceedings of the 2022 {Conference} on {Empirical} {Methods} in {Natural} {Language} {Processing}},
	publisher = {Association for Computational Linguistics},
	author = {Ahuja, Kabir and Sitaram, Sunayana and Dandapat, Sandipan and Choudhury, Monojit},
	month = dec,
	year = {2022},
	pages = {4310--4323},
}

@inproceedings{naeini2015obtaining,
  title={Obtaining well calibrated probabilities using bayesian binning},
  author={Naeini, Mahdi Pakdaman and Cooper, Gregory and Hauskrecht, Milos},
  booktitle={Proceedings of the AAAI conference on artificial intelligence},
  volume={29},
  number={1},
  year={2015}
}

@inproceedings{Zhang2015CharacterlevelCN,
  title={Character-level Convolutional Networks for Text Classification},
  author={Xiang Zhang and Junbo Jake Zhao and Yann LeCun},
  booktitle={NIPS},
  year={2015}
}

@inproceedings{zadrozny2001obtaining,
  title={Obtaining calibrated probability estimates from decision trees and naive bayesian classifiers},
  author={Zadrozny, Bianca and Elkan, Charles},
  booktitle={Icml},
  volume={1},
  pages={609--616},
  year={2001}
}

@article{platt1999probabilistic,
  title={Probabilistic outputs for support vector machines and comparisons to regularized likelihood methods},
  author={Platt, John and others},
  journal={Advances in large margin classifiers},
  volume={10},
  number={3},
  pages={61--74},
  year={1999},
  publisher={Cambridge, MA}
}

@inproceedings{wang-culotta-2020-identifying,
    title = "Identifying Spurious Correlations for Robust Text Classification",
    author = "Wang, Zhao  and
      Culotta, Aron",
    booktitle = "Findings of the Association for Computational Linguistics: EMNLP 2020",
    month = nov,
    year = "2020",
    address = "Online",
    publisher = "Association for Computational Linguistics",
    url = "https://aclanthology.org/2020.findings-emnlp.308",
    doi = "10.18653/v1/2020.findings-emnlp.308",
    pages = "3431--3440",
}

@inproceedings{wang-etal-2022-identifying,
    title = "Identifying and Mitigating Spurious Correlations for Improving Robustness in {NLP} Models",
    author = "Wang, Tianlu  and
      Sridhar, Rohit  and
      Yang, Diyi  and
      Wang, Xuezhi",
    booktitle = "Findings of the Association for Computational Linguistics: NAACL 2022",
    month = jul,
    year = "2022",
    address = "Seattle, United States",
    publisher = "Association for Computational Linguistics",
    url = "https://aclanthology.org/2022.findings-naacl.130",
    doi = "10.18653/v1/2022.findings-naacl.130",
    pages = "1719--1729",
}

@inproceedings{sap-etal-2020-commonsense,
    title = "Commonsense Reasoning for Natural Language Processing",
    author = "Sap, Maarten  and
      Shwartz, Vered  and
      Bosselut, Antoine  and
      Choi, Yejin  and
      Roth, Dan",
    booktitle = "Proceedings of the 58th Annual Meeting of the Association for Computational Linguistics: Tutorial Abstracts",
    month = jul,
    year = "2020",
    address = "Online",
    publisher = "Association for Computational Linguistics",
    url = "https://aclanthology.org/2020.acl-tutorials.7",
    doi = "10.18653/v1/2020.acl-tutorials.7",
    pages = "27--33",
}

@article{peters2018deep,
  title={Deep contextualized word representations. arXiv 2018},
  author={Peters, ME and Neumann, M and Iyyer, M and Gardner, M and Clark, C and Lee, K and Zettlemoyer, L},
  journal={arXiv preprint arXiv:1802.05365},
  volume={12},
  year={2018}
}

@misc{chen_bert_2019,
	title = {{BERT} for {Joint} {Intent} {Classification} and {Slot} {Filling}},
	url = {http://arxiv.org/abs/1902.10909},
	urldate = {2022-12-22},
	publisher = {arXiv},
	author = {Chen, Qian and Zhuo, Zhu and Wang, Wen},
	month = feb,
	year = {2019},
	note = {arXiv:1902.10909 [cs]},
}

@misc{alt_fine-tuning_2019,
	title = {Fine-tuning {Pre}-{Trained} {Transformer} {Language} {Models} to {Distantly} {Supervised} {Relation} {Extraction}},
	url = {http://arxiv.org/abs/1906.08646},
	doi = {10.48550/arXiv.1906.08646},
	urldate = {2022-12-22},
	publisher = {arXiv},
	author = {Alt, Christoph and Hübner, Marc and Hennig, Leonhard},
	month = jun,
	year = {2019},
	note = {arXiv:1906.08646 [cs]},
}

@inproceedings{guo_calibration_2017,
	title = {On {Calibration} of {Modern} {Neural} {Networks}},
	url = {https://proceedings.mlr.press/v70/guo17a.html},
	language = {en},
	urldate = {2023-03-20},
	booktitle = {Proceedings of the 34th {International} {Conference} on {Machine} {Learning}},
	publisher = {PMLR},
	author = {Guo, Chuan and Pleiss, Geoff and Sun, Yu and Weinberger, Kilian Q.},
	month = jul,
	year = {2017},
	note = {ISSN: 2640-3498},
	pages = {1321--1330},
}

@inproceedings{devlin_bert_2019,
	address = {Minneapolis, Minnesota},
	title = {{BERT}: {Pre}-training of {Deep} {Bidirectional} {Transformers} for {Language} {Understanding}},
	shorttitle = {{BERT}},
	url = {https://aclanthology.org/N19-1423},
	doi = {10.18653/v1/N19-1423},
	urldate = {2023-03-20},
	booktitle = {Proceedings of the 2019 {Conference} of the {North} {American} {Chapter} of the {Association} for {Computational} {Linguistics}: {Human} {Language} {Technologies}, {Volume} 1 ({Long} and {Short} {Papers})},
	publisher = {Association for Computational Linguistics},
	author = {Devlin, Jacob and Chang, Ming-Wei and Lee, Kenton and Toutanova, Kristina},
	month = jun,
	year = {2019},
	pages = {4171--4186},
}

@misc{desai_calibration_2020,
	title = {Calibration of {Pre}-trained {Transformers}},
	url = {http://arxiv.org/abs/2003.07892},
	urldate = {2023-04-03},
	publisher = {arXiv},
	author = {Desai, Shrey and Durrett, Greg},
	month = oct,
	year = {2020},
	note = {arXiv:2003.07892 [cs]},
}

@misc{kim_bag_2023,
	title = {Bag of {Tricks} for {In}-{Distribution} {Calibration} of {Pretrained} {Transformers}},
	url = {http://arxiv.org/abs/2302.06690},
	urldate = {2023-04-03},
	publisher = {arXiv},
	author = {Kim, Jaeyoung and Na, Dongbin and Choi, Sungchul and Lim, Sungbin},
	month = feb,
	year = {2023},
	note = {arXiv:2302.06690 [cs]},
}

@article{moon_masker_2021,
	title = {{MASKER}: {Masked} {Keyword} {Regularization} for {Reliable} {Text} {Classification}},
	volume = {35},
	copyright = {Copyright (c) 2021 Association for the Advancement of Artificial Intelligence},
	issn = {2374-3468},
	shorttitle = {{MASKER}},
	url = {https://ojs.aaai.org/index.php/AAAI/article/view/17601},
	doi = {10.1609/aaai.v35i15.17601},
	language = {en},
	number = {15},
	urldate = {2023-04-03},
	journal = {Proceedings of the AAAI Conference on Artificial Intelligence},
	author = {Moon, Seung Jun and Mo, Sangwoo and Lee, Kimin and Lee, Jaeho and Shin, Jinwoo},
	month = may,
	year = {2021},
	note = {Number: 15},
	pages = {13578--13586},
}

@inproceedings{hendrycks_pretrained_2020,
	address = {Online},
	title = {Pretrained {Transformers} {Improve} {Out}-of-{Distribution} {Robustness}},
	url = {https://aclanthology.org/2020.acl-main.244},
	doi = {10.18653/v1/2020.acl-main.244},
	urldate = {2023-04-04},
	booktitle = {Proceedings of the 58th {Annual} {Meeting} of the {Association} for {Computational} {Linguistics}},
	publisher = {Association for Computational Linguistics},
	author = {Hendrycks, Dan and Liu, Xiaoyuan and Wallace, Eric and Dziedzic, Adam and Krishnan, Rishabh and Song, Dawn},
	month = jul,
	year = {2020},
	pages = {2744--2751},
}

@article{kong_calibrated_2020,
	title = {Calibrated {Language} {Model} {Fine}-{Tuning} for {In}- and {Out}-of-{Distribution} {Data}},
	url = {https://www.semanticscholar.org/reader/05ef3566499e24888a8c500944336616f8f418a4},
	doi = {10.18653/v1/2020.emnlp-main.102},
	language = {en},
	urldate = {2023-04-04},
	journal = {ArXiv},
	author = {Kong, Lingkai and Jiang, Haoming and Zhuang, Yuchen and Lyu, Jie and Zhao, T. and Zhang, Chao},
	year = {2020},
}

@inproceedings{dan_effects_2021,
	address = {Punta Cana, Dominican Republic},
	title = {On the {Effects} of {Transformer} {Size} on {In}- and {Out}-of-{Domain} {Calibration}},
	url = {https://aclanthology.org/2021.findings-emnlp.180},
	doi = {10.18653/v1/2021.findings-emnlp.180},
	urldate = {2023-04-04},
	booktitle = {Findings of the {Association} for {Computational} {Linguistics}: {EMNLP} 2021},
	publisher = {Association for Computational Linguistics},
	author = {Dan, Soham and Roth, Dan},
	month = nov,
	year = {2021},
	pages = {2096--2101},
}

@misc{chen_close_2022,
	title = {A {Close} {Look} into the {Calibration} of {Pre}-trained {Language} {Models}},
	url = {http://arxiv.org/abs/2211.00151},
	urldate = {2023-04-04},
	publisher = {arXiv},
	author = {Chen, Yangyi and Yuan, Lifan and Cui, Ganqu and Liu, Zhiyuan and Ji, Heng},
	month = nov,
	year = {2022},
	note = {arXiv:2211.00151 [cs]},
}

@inproceedings{thulasidasan_mixup_2019,
	title = {On {Mixup} {Training}: {Improved} {Calibration} and {Predictive} {Uncertainty} for {Deep} {Neural} {Networks}},
	volume = {32},
	shorttitle = {On {Mixup} {Training}},
	url = {https://proceedings.neurips.cc/paper/2019/hash/36ad8b5f42db492827016448975cc22d-Abstract.html},
	urldate = {2023-04-04},
	booktitle = {Advances in {Neural} {Information} {Processing} {Systems}},
	publisher = {Curran Associates, Inc.},
	author = {Thulasidasan, Sunil and Chennupati, Gopinath and Bilmes, Jeff A and Bhattacharya, Tanmoy and Michalak, Sarah},
	year = {2019},
}

@misc{malinin_predictive_2018,
	title = {Predictive {Uncertainty} {Estimation} via {Prior} {Networks}},
	url = {http://arxiv.org/abs/1802.10501},
	doi = {10.48550/arXiv.1802.10501},
	urldate = {2023-04-04},
	publisher = {arXiv},
	author = {Malinin, Andrey and Gales, Mark},
	month = nov,
	year = {2018},
	note = {arXiv:1802.10501 [cs, stat]},
}

@inproceedings{du-etal-2021-towards,
    title = "Towards Interpreting and Mitigating Shortcut Learning Behavior of {NLU} models",
    author = "Du, Mengnan  and
      Manjunatha, Varun  and
      Jain, Rajiv  and
      Deshpande, Ruchi  and
      Dernoncourt, Franck  and
      Gu, Jiuxiang  and
      Sun, Tong  and
      Hu, Xia",
    booktitle = "Proceedings of the 2021 Conference of the North American Chapter of the Association for Computational Linguistics: Human Language Technologies",
    month = jun,
    year = "2021",
    address = "Online",
    publisher = "Association for Computational Linguistics",
    url = "https://aclanthology.org/2021.naacl-main.71",
    doi = "10.18653/v1/2021.naacl-main.71",
    pages = "915--929",
}

@inproceedings{barbieri2020,
    title = "{T}weet{E}val: Unified Benchmark and Comparative Evaluation for Tweet Classification",
    author = "Barbieri, Francesco  and
      Camacho-Collados, Jose  and
      Espinosa Anke, Luis  and
      Neves, Leonardo",
    booktitle = "Findings of the Association for Computational Linguistics: EMNLP 2020",
    month = nov,
    year = "2020",
    address = "Online",
    publisher = "Association for Computational Linguistics",
    url = "https://aclanthology.org/2020.findings-emnlp.148",
    doi = "10.18653/v1/2020.findings-emnlp.148",
    pages = "1644--1650",
}

@inproceedings{hovy-etal-2001-toward,
    title = "Toward Semantics-Based Answer Pinpointing",
    author = "Hovy, Eduard  and
      Gerber, Laurie  and
      Hermjakob, Ulf  and
      Lin, Chin-Yew  and
      Ravichandran, Deepak",
    booktitle = "Proceedings of the First International Conference on Human Language Technology Research",
    year = "2001",
    url = "https://www.aclweb.org/anthology/H01-1069",
}

@article{warstadt2019neural,
  title={Neural Network Acceptability Judgments},
  author={Warstadt, Alex and Singh, Amanpreet and Bowman, Samuel},
  journal={Transactions of the Association for Computational Linguistics},
  volume={7},
  pages={625--641},
  year={2019}
}

@inproceedings{socher2013,
    title = "Recursive Deep Models for Semantic Compositionality Over a Sentiment Treebank",
    author = "Socher, Richard  and
      Perelygin, Alex  and
      Wu, Jean  and
      Chuang, Jason  and
      Manning, Christopher D.  and
      Ng, Andrew  and
      Potts, Christopher",
    booktitle = "Proceedings of the 2013 Conference on Empirical Methods in Natural Language Processing",
    month = oct,
    year = "2013",
    address = "Seattle, Washington, USA",
    publisher = "Association for Computational Linguistics",
    url = "https://aclanthology.org/D13-1170",
    pages = "1631--1642",
}

@inproceedings{lanalbert,
  title={ALBERT: A Lite BERT for Self-supervised Learning of Language Representations},
  author={Lan, Zhenzhong and Chen, Mingda and Goodman, Sebastian and Gimpel, Kevin and Sharma, Piyush and Soricut, Radu},
  booktitle={International Conference on Learning Representations}
}

@inproceedings{hedeberta,
  title={DEBERTA: DECODING-ENHANCED BERT WITH DISENTANGLED ATTENTION},
  author={He, Pengcheng and Liu, Xiaodong and Gao, Jianfeng and Chen, Weizhu},
  booktitle={International Conference on Learning Representations}
}

@inproceedings{sundararajan2017axiomatic,
  title={Axiomatic attribution for deep networks},
  author={Sundararajan, Mukund and Taly, Ankur and Yan, Qiqi},
  booktitle={International conference on machine learning},
  pages={3319--3328},
  year={2017},
  organization={PMLR}
}

@inproceedings{lewis2020bart,
  title={BART: Denoising Sequence-to-Sequence Pre-training for Natural Language Generation, Translation, and Comprehension},
  author={Lewis, Mike and Liu, Yinhan and Goyal, Naman and Ghazvininejad, Marjan and Mohamed, Abdelrahman and Levy, Omer and Stoyanov, Veselin and Zettlemoyer, Luke},
  booktitle={Proceedings of the 58th Annual Meeting of the Association for Computational Linguistics},
  pages={7871--7880},
  year={2020}
}

\end{document}